%% file: 0.Main.tex
\documentclass{article}

\usepackage{microtype}
\usepackage{graphicx}
\usepackage{subfigure}
\usepackage{booktabs} 
\usepackage{amsmath}
\usepackage{amsfonts}
\usepackage{subcaption}
\usepackage{grffile}
\usepackage{booktabs}
\usepackage{tabularx}

\usepackage{hyperref}

\usepackage{format}

\mlsystitlerunning{Style Extraction on Text Embeddings via VAE}

\begin{document}

\twocolumn[
\mlsystitle{Style Extraction on Text Embeddings\\ Using VAE and Parallel Dataset}

\vspace{-1.5em}
\begin{center}
    \textbf{September 2, 2024}
\end{center}

\begin{mlsysauthorlist}
\mlsysauthor{InJin Kong}{snu}
\mlsysauthor{Shinyee Kang}{snu}
\mlsysauthor{Yuna Park}{snu}
\mlsysauthor{Sooyong Kim}{snu}
\mlsysauthor{Sanghyun Park}{snu}
\end{mlsysauthorlist}

\mlsysaffiliation{snu}{Seoul National University}


\mlsyssetsymbol{equal}

\vskip 0.3in

\begin{abstract}
This study investigates the stylistic differences among various Bible translations using a Variational Autoencoder (VAE) model. By embedding textual data into high-dimensional vectors, the study aims to detect and analyze stylistic variations between translations, with a specific focus on distinguishing the American Standard Version (ASV) from other translations. The results demonstrate that each translation exhibits a unique stylistic distribution, which can be effectively identified using the VAE model. These findings suggest that the VAE model is proficient in capturing and differentiating textual styles, although it is primarily optimized for distinguishing a single style. The study highlights the model's potential for broader applications in AI-based text generation and stylistic analysis, while also acknowledging the need for further model refinement to address the complexity of multi-dimensional stylistic relationships. Future research could extend this methodology to other text domains, offering deeper insights into the stylistic features embedded within various types of textual data.
\end{abstract}]

\input{1.Introduction}

\input{2.Related_Work}

\input{3.Methodology}

\input{4.Results_and_Discussion}

\input{5.Conclusion}

\bibliography{citation}
\bibliographystyle{format}

\end{document}

%% file: 1.Introduction.tex
\section{Introduction}
\label{sec:intro}

Language in both speech and writing consists of two essential components: content and style. Broadly speaking, content refers to what is being expressed, while style pertains to how it is expressed. Specifically, style encompasses the variability of linguistic forms in actual language use \cite{inbook}. Historically, "Style" has been explored within the field of Stylistics—a branch of applied linguistics that examines writing styles in literary criticism as well as tone and accent in discourse analysis. However, the notion of style has gained renewed attention with the advancement of Generative AI technologies. Since the seminal work of \cite{gatys2015neuralalgorithmartisticstyle}, which introduced style transfer technology for images, substantial progress has been made in the field of computer vision \cite{gatys2015neuralalgorithmartisticstyle}. This progress has extended to text-style transfer, leading to novel applications in AI-driven text generation, such as Character AI.

Textual style, however, presents a more subjective and nuanced challenge compared to visual styles due to the complex interplay of various linguistic elements \cite{corpus}. Unlike paraphrasing, which focuses on semantic similarity and lexical variation \cite{shen2022evaluationmetricsparaphrasegeneration}, text-style involves additional dimensions such as syntactic structures (e.g., active vs. passive voice, tense) and thematic elements (e.g., emphasis on certain parts of speech such as verbs or adjectives) \cite{lyu2021styleptbcompositionalbenchmarkfinegrained}. These complexities raise challenges in evaluating text-style transfer. Conventional evaluation metrics like BLEU, ROUGE, BERTScore, and perplexity, often based on n-gram overlaps, fail to capture the stylistic nuances of text and can be ineffective in this context. Although human evaluation offers a more accurate assessment, it is typically impractical due to its time-consuming and costly nature.

One of the strengths of large language models (LLMs) is their ability to represent words as high-dimensional vectors that capture semantic relationships more effectively than traditional n-gram-based models \cite{mikolov-etal-2013-linguistic}. While n-gram models treat words as discrete and unrelated units, LLMs utilize continuous embedding spaces, where vectors for semantically related words are closer together. This allows LLMs to better handle the complexity of style beyond simple lexical variations. The linear representation hypothesis suggests that styles in text can be understood as linear transformations within the embedding space \cite{mikolov2013distributedrepresentationswordsphrases}. 

To address the limitations of current evaluation methods, we propose a geometric approach to evaluating text-style by leveraging the vector representations of texts in embedding spaces. By analyzing the transformations within these spaces, we aim to provide a more precise and automated method for assessing stylistic differences in text. This approach offers the potential for more robust and scalable evaluations of text-style transfer, without relying solely on human judgment.

Our findings indicate that this geometric method not only aligns well with human evaluations but also significantly reduces the overhead involved in the evaluation process. The implications of this work are far-reaching, offering potential applications in automated content creation, personalized writing assistants, and AI-driven literary analysis tools.

%% file: 2.Related_Work.tex
\section{Related Works}\label{sec:related_works}

\subsection{Styles in Natural Languages}
In linguistics, style has been seen as the unique way individuals or groups engage in conversation, conveying politeness or formality, and able to be controlled and adjusted to suit the intended social context \cite{Labov1997}. Additionally, style is the set of linguistic features such as tone, punctuation, word choice, and syntactic structure, playing a key role in stylistics and sentiment analysis \cite{pang}. 

Subsequently, research on style transfer in text has considered style as the specific manner in which ideas are expressed in text, distinguishable from the content \cite{shen}. In style transfer tasks, style is represented by personal style, formality, politeness, offensiveness, genre, and sentiment \cite{Toshevska_2022}. Current studies of styles focus on computational models for style transfer; Cross-Alignment with non-parallel text \cite{shen}, Retrieve-and-Edit approach \cite{li-etal-2018-delete}, Unsupervised style transfer \cite{prabhumoye-etal-2018-style}, Generative probabilistic model \cite{he2020probabilisticformulationunsupervisedtext}, and Reinforcement Learning for style transfer \cite{gong2019reinforcementlearningbasedtext}.

\subsection{Evaluating Style Transfer in Text}
Evaluation metrics are vital for text style transfer as they provide precise, quantitative assessments of how effectively the generated text adheres to the target stylistic attributes while preserving semantic integrity. However, evaluation can be challenging due to the subjective nature of style. It typically involves automatic evaluation and human evaluation \cite{jin-etal-2022-deep}.

\subsubsection{Automatic Evaluation}
The automatic evaluation measures how well the meaning of the original sentence was preserved in the output (generated sentence). The following metrics are commonly used:
\begin{itemize}
    \item BLEU: Measures n-gram precision between generated text and references \cite{papineni-etal-2002-bleu}.
    \item ROUGE: Assesses overlap of n-grams, focusing on recall to evaluate content coverage \cite{lin-2004-rouge}.
    \item METEOR: Evaluates translation quality using precision, recall, stemming, and synonymy \cite{banerjee-lavie-2005-meteor}.
    \item BERTScore: Utilizes BERT embeddings to measure semantic similarity between generated and reference texts \cite{zhang2020bertscoreevaluatingtextgeneration}.
\end{itemize}

\subsubsection{Human Evaluation}
Human judges assess how well the generated text adheres to the desired style and maintains semantic integrity. \cite{Yamshchikov_2021} delineates the distinctions between human evaluation and automatic methods, illustrating how human assessment captures nuanced stylistic and semantic subtleties. However, it is costly and lacks the consistency, objectivity, and scalability provided by automatic evaluation methods. Additionally, both methods are limited by their reliance on reference texts, which may not fully capture the breadth of acceptable outputs or the creative potential of the generated text. 

\subsection{Anomaly Detection and VAE}
Anomaly detection has evolved through various methodologies to address the challenge of identifying outliers across different domains \cite{NIPS1999_8725fb77, 4781136}. The advent of deep learning introduced Autoencoders \cite{article}, which makes it possible to detect anomalies in high-dimensional data by analyzing reconstruction errors. Further advancements have been made with Variational Autoencoders (VAE), which leverage both probabilistic modeling and latent space representations \cite{kingma2022autoencodingvariationalbayes}. We will employ a VAE model, trained on high-dimensional embedding vectors representing a single stylistic attribute, to identify anomalies by capturing deviations in stylistic characteristics.

%% file: 3.Methodology.tex
\section{Methodology}\label{sec:methodology}

\subsection{Data Collection and Preprocessing}
This study utilizes biblical data collected from \textit{Bible SuperSearch} \cite{biblesupersearch}, a platform operating under the GNU GPL open source license. Ten different versions were initially considered: KJV, NET, ASV, ASVS, Coverdale, Geneva, KJV\_Strongs, Bishops, Tyndale, and WEB. However, Bishops, Tyndale, and WEB were excluded due to insufficient parallel data. The remaining versions were selected for their linguistic diversity and historical backgrounds to enhance the depth of our style classification study.

The biblical texts are publicly available under the GNU GPL license, allowing free use for research purposes. Our study adhered to these guidelines without altering the original texts. In the preprocessing phase, we extracted the biblical data in JSON format and encoded all text files using UTF-8 to handle special characters. The initial data quality was high, minimizing the need for extensive text cleaning.

\subsection{Embedding and Model Training}
We employed OpenAI's text-embedding-3-small model to embed each biblical sentence into 1536-dimensional vectors. This model was chosen for its balance between performance and computational efficiency, making it suitable for our research. These high-dimensional vectors capture the nuanced language style of the sentences, providing foundational data for style-based classification.

\subsection{Style Extraction}\label{subsec:style_extraction}
Text embedding is assumed to include both content and style, as represented by the following equation:

\[
\text{text\_embedding} = \text{style\_embedding} + \text{content\_embedding}
\]

Under this assumption, text embedding can be seen as simultaneously containing both the content and stylistic features of the text. In this study, we utilized this assumption to perform an analysis based on Bible data. The Bible data consists of the same verse expressed in multiple translations in a parallel structure, where the content remains the same, but the style varies. This characteristic of Bible data justifies the assumption that each translation’s content embedding is identical. That is, the differences between the translations are primarily due to style, allowing for style analysis to be conducted. The core assumption of this study is that the difference in text embeddings between translations reflects the difference in styles. This can be expressed mathematically as follows:

\begin{multline*}
\text{KJV\_embedding} - \text{Other\_embedding} = \\
\text{KJV\_style\_embedding} - \text{Other\_style\_embedding}
\end{multline*}

Through this relationship, we calculated the difference between the two text embeddings and, based on this, measured the difference in style between the translations. Specifically, the goal of the study was to analyze the text embedding differences between KJV (King James Version) and other translations (e.g., ASV (American Standard Version)) to quantify the stylistic features. To do this, we calculated the difference between embeddings, represented as 1536-dimensional vectors, and used Variational Autoencoder (VAE) as a tool to analyze the distribution of these vectors.

The VAE is an unsupervised learning method that models the distribution of data in a latent space. In this study, we aimed to utilize the VAE to classify the embedding differences between translations and detect stylistic differences through anomaly detection. By compressing the input data and reconstructing it, VAE retains the important features while learning the distribution, allowing for the modeling of stylistic differences between translations.

During the training process of the VAE, we used the distribution differences between \textit{KJV\_embedding} and \textit{ASV\_embedding}. The VAE learned the difference between KJV and ASV embeddings in the latent space and then measured the similarity between the reconstructed distribution and the original distribution. We computed the L2-norm in this reconstruction process to quantitatively evaluate the stylistic similarity or difference between the translations. This allowed us to analyze the stylistic differences between KJV and ASV, as well as conduct comparative analyses with other translations.

\begin{table}[htbp]
\caption{Notation used throughout this article.}
\label{tab:notation}
\vskip 0.15in
\begin{center}
\begin{small}
\begin{sc}
\begin{tabular}{p{0.12\textwidth} p{0.25\textwidth}}
\toprule
\textbf{Symbol} & \textbf{Description} \\
\midrule
$\mathbf{k}^{(i)}$ & Embedding of KJV, \newline $i = 1, \cdots, N$ \\
$\mathbf{a}^{(i)}$ & Embedding of ASV \\
$\mathbf{y}^{(i)}_j$ & Embedding of other Bibles,\newline $j = 1, \cdots, 5$ \\
$\mathbf{x}^{(i)}$ & KJV\_style\_embedding \newline $-$ ASV\_style\_embedding \\
$\mathbb{R}^d$ & $d$-dimensional input space \\
$\mathbb{R}^p$ & $p$-dimensional \newline feature space ($p < d$) \\
$\psi:\mathbb{R}^d \rightarrow \mathbb{R}^p$ & Encoder of VAE \\
$\theta:\mathbb{R}^p \rightarrow \mathbb{R}^d$ & Decoder of VAE \\
\bottomrule
\end{tabular}
\end{sc}
\end{small}
\end{center}
\vskip -0.1in
\end{table}

\begin{algorithm}[htbp]
   \caption{VAE Training Process}
   \label{alg:vae_training}
\begin{algorithmic}
   \STATE {\bfseries Input:} $\mathbf{x}^{(i)}$ \quad \COMMENT{Training data}
   \STATE {\bfseries Output:} $w_{\psi}$ (encoder parameters), $w_{\theta}$ (decoder parameters)
   \STATE Initialize parameters $w_{\psi}$, $w_{\theta}$
   \REPEAT
      \FOR{$i=1$ {\bfseries to} $N$}
         \STATE $z^{(i)} = \psi(\mathbf{x}^{(i)}, w_{\psi})$ \quad \COMMENT{Generate feature vectors}
         \STATE $\hat{\mathbf{x}}^{(i)} = \theta(z^{(i)}, w_{\theta})$ \quad \COMMENT{Reconstruct original embedding}
      \ENDFOR
      \STATE $\mathcal{L}_{\text{mse}} = \frac{1}{N} \sum_{i=1}^{N} (\mathbf{x}^{(i)} - \hat{\mathbf{x}}^{(i)})^2$
      \STATE Update $w_{\psi}$ and $w_{\theta}$ using gradients of $\mathcal{L}_{\text{mse}}$
   \UNTIL{parameters $w_{\psi}$ and $w_{\theta}$ converge}
   \STATE \textbf{return} $w_{\psi}$, $w_{\theta}$
\end{algorithmic}
\end{algorithm}

 In conclusion, this study evaluated the stylistic differences between Bible translations using VAE for anomaly detection. Through this process, we effectively quantified the stylistic similarities and differences between various translations. Based on the VAE model, trained on the difference between \textit{KJV\_embedding} and \textit{ASV\_embedding}, we similarly analyzed the stylistic differences between other translations. This methodology enabled sophisticated text analysis that went beyond merely examining content features to include stylistic features. Thus, we provided new insights into how stylistic differences manifest within the embedding space

\subsection{Model Architecture and Training Details}
The VAE model used in this study has an input dimension of 1536, and both encoder and decoder use fully connected (FC) layers. The size of each hidden layer follows a geometric sequence from the input dimension of 1536 to the final feature dimension (rounded to the nearest integer). Batch normalization is applied to all layers except the final output layers of both the encoder and decoder. The activation function used is Leaky ReLU ($\alpha$=1e-2) except for the final output layer of the encoder and decoder. The final output layer of the decoder uses a Sigmoid-based activation function to ensure that the output distribution lies within the range [-1,1].

The hyperparameters are as follows: 6 values for the number of hidden layers (ranging from 1 to 6) and 6 values for the feature dimension (ranging from $2^3$ to $2^8$), resulting in 36 total combinations.

\begin{figure}[htbp]
    \centering
    \includegraphics[width=0.5\textwidth]{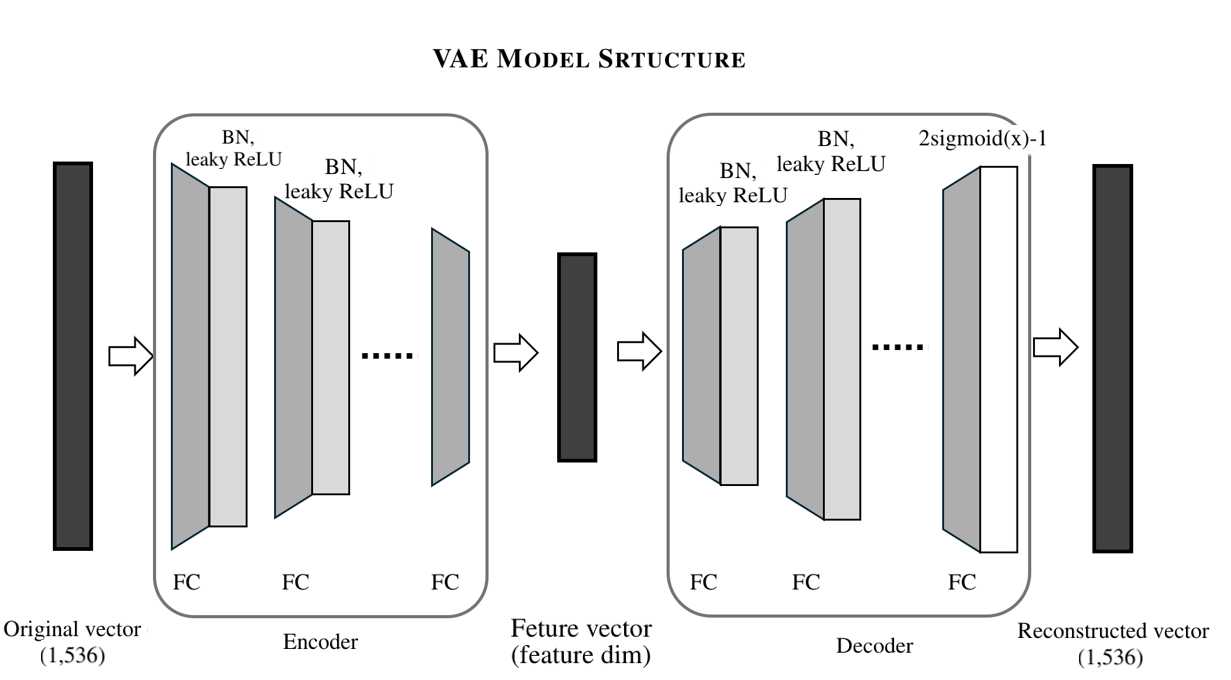}
    \caption{A schematic illustration of the VAE model. The encoder receives a 1,536-dimensional original (sentence embedding) vector as input and outputs a feature vector of the feature dimension. The decoder takes the feature vector of the feature dimension as input and outputs a 1,536-dimensional reconstructed vector.}
    \label{fig:model structure}
\end{figure}

We split 13,823 sentence vectors into training and test sets with a 9:1 ratio, using KJV-ASV differences as training data. The model employs fully connected layers with batch normalization and Leaky ReLU activation, and is trained using the Adam optimizer and MSE loss function. A schematic of the model structure is provided in Figure \ref{fig:model structure}.

\subsection{Evaluation Metrics}\label{subsec:evaluation_metrics}
According to our hypothesis, the KJV-ASV vector is expected to contain information related to the style of ASV, with KJV as the reference point. If a VAE with a sufficiently small feature dimension can effectively reconstruct this vector, it suggests that the VAE is leveraging specific stylistic features during the encoding-decoding process. On the other hand, if data not included in the model’s training process are reconstructed through the VAE, the reconstruction quality is expected to be poor compared to the original. Based on this characteristic, we aim to perform anomaly detection using the VAE.

We aim to verify whether the VAE, trained using KJV-ASV vectors, has effectively learned the unique style of ASV. To do so, the trained VAE will be applied to six Bible translations (ASV, NET, ASVS, Coverdale, Geneva, and KJV Strongs), and we will examine if the model successfully distinguishes ASV’s unique style compared to other translations. For the test dataset (not used during model training), ASV will serve as the normal data, and the other five translations (NET, ASVS, Coverdale, Geneva, and KJV Strongs) will serve as anomaly data, consisting of sentence embedding vectors corresponding to the same Bible verses as in the test dataset. To remove the context of KJV during ASV training, the VAE was trained on the differences between the sentence vectors of ASV and KJV (KJV-ASV). Similarly, the anomaly data from the other Bible translations will be processed by subtracting the corresponding KJV sentence vectors, following the same procedure.

Among the 36 hyperparameter sets, the model that most clearly differentiates the reconstruction L2 error distribution between the training data and the anomalies will be considered the most effective in detecting the unique style of ASV. We will evaluate how well the original data and anomaly data are distinguished using Fisher’s Linear Discriminant (FLD). FLD increases as the squared difference between the means of the two distributions becomes larger, and the sum of their variances becomes smaller. The formula for FLD \( S \) is as follows:

\[
S = \frac{(\mu_1 - \mu_2)^2}{\sigma_1^2 + \sigma_2^2}
\]

where \( \mu_1 \) and \( \mu_2 \) are the means of the original data and anomaly data distributions, respectively, and \( \sigma_1 \) and \( \sigma_2 \) are the variances of the original data and anomaly data distributions, respectively. This metric will help quantify how well the model separates the 
original data from anomalies based on reconstruction errors.

\begin{algorithm}[htbp]
   \caption{Anomaly Detection by VAE}
   \label{alg:anomaly_detection}
\begin{algorithmic}[1]
   \STATE \textbf{Input:} $\mathbf{a}^{(i)}, \mathbf{y}^{(i)}_j$, Trained parameters by $\mathbf{x}^{(i)}$: $w_{\psi}, w_{\theta}$, $\alpha = 0.1, \ldots, 1.4$
   \STATE \textbf{Output:} Fisher’s Linear Discriminant (FLD): $S_j$
   \FOR{$i = 1$ \textbf{to} $N$}
      \STATE $z^{(i)} = \psi(\mathbf{a}^{(i)}, w_{\psi})$
      \STATE $\hat{\mathbf{a}}^{(i)} = \theta(z^{(i)}, w_{\theta})$
      \FOR{$j = 1$ \textbf{to} $5$}
         \STATE $z^{(i)}_j = \psi(\mathbf{y}^{(i)}_j, w_{\psi})$
         \STATE $\hat{\mathbf{y}}^{(i)}_j = \theta(z^{(i)}_j, w_{\theta})$
      \ENDFOR
   \ENDFOR
   \STATE $\ell_{2,\mathbf{a}} = \|\mathbf{a}^{(i)} - \hat{\mathbf{a}}^{(i)}\|_2$
   \STATE $\ell_{2,\mathbf{y}_j} = \|\mathbf{y}^{(i)}_j - \hat{\mathbf{y}}^{(i)}_j\|_2$
   \STATE $\mu_{\mathbf{a}}, \sigma_{\mathbf{a}}, \mu_{\mathbf{y}_j}, \sigma_{\mathbf{y}_j} \leftarrow$ mean and standard deviation of $\ell_{2,\mathbf{a}}, \ell_{2,\mathbf{y}_j}$
   \STATE Find the threshold minimizing total error: $\gamma = \mu_{\mathbf{a}} + \alpha \sigma_{\mathbf{a}}$
   \STATE $S_j = \frac{(\mu_{\mathbf{a}} - \mu_{\mathbf{y}_j})^2}{\sigma_{\mathbf{a}}^2 + \sigma_{\mathbf{y}_j}^2}$
   \IF{$S_j > \gamma$}
      \STATE $\mathbf{y}_j$ is anomaly
   \ELSE
      \STATE $\mathbf{y}_j$ is not anomaly
   \ENDIF
   \STATE \textbf{return} $S_j$
\end{algorithmic}
\end{algorithm}

%% file: 4.Results_and_Discussion.tex
\section{Results}\label{sec:results}

\subsection{Training Convergence and Loss Analysis}

\begin{figure}[htbp]
    \centering
    \includegraphics[width=0.5\textwidth]{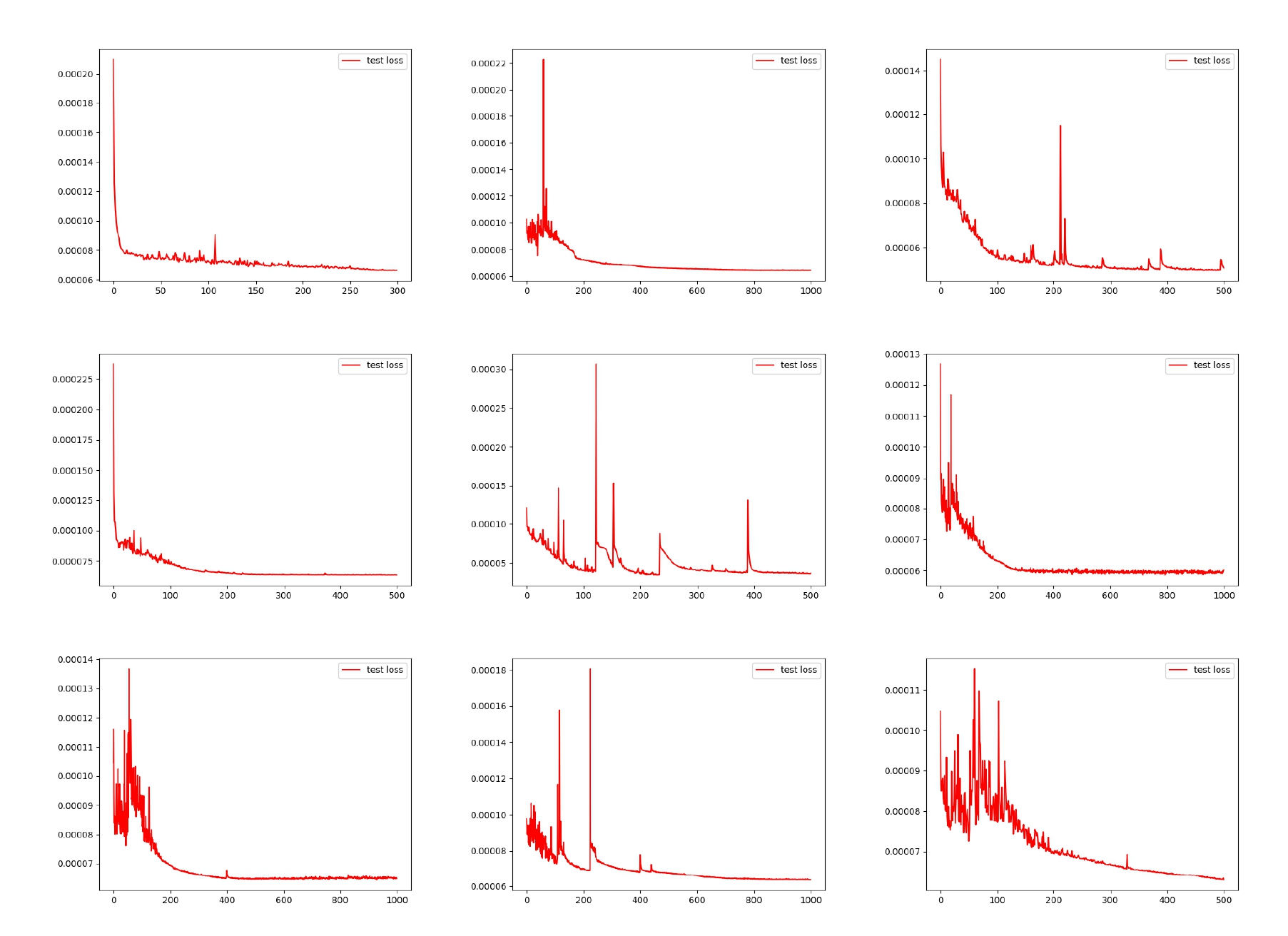}
    \caption{Test set loss during training. The x-axis represents the number of epochs, and the y-axis represents the mean error. The hyperparameters of each model are as follows: starting from left the 1st, 2nd, and 3rd columns represent feature dimensions of 8, 64, and 256, respectively, and the starting from top 1st, 2nd, and 3rd rows represent 1, 3, and 6 hidden layers, respectively.}
    \label{fig:appendix_loss}
\end{figure}

For all 36 hyperparameter combinations, both the training loss and test loss decreased and eventually converged, indicating that the models successfully learned from the data and reached a stable state in terms of reconstruction error. Detailed loss curves and analysis are provided in Figure \ref{fig:appendix_loss}.

\subsection{L2 Error Distribution and FLD Analysis}

\begin{figure}[htbp]
    \centering
    \includegraphics[width=0.5\textwidth]{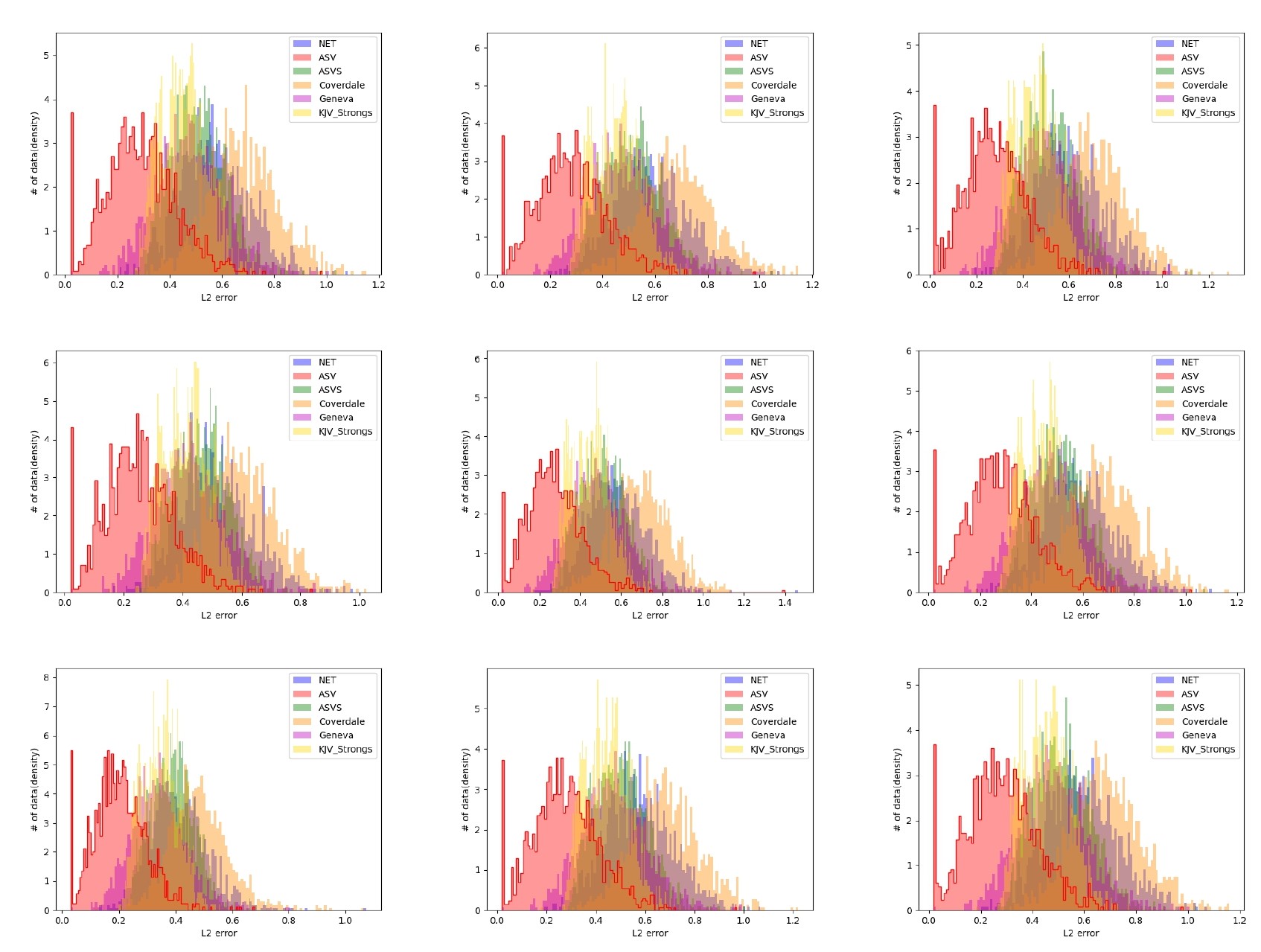}
    \caption{L2 error distribution on ASV, NET, ASVS, Coverdale, Geneva, and KJV Strongs. The x-axis represents the L2 error between the original and reconstructed sentence vector, and the y-axis represents the distribution density. The hyperparameters of each model are as follows: starting from left the 1st, 2nd, and 3rd columns represent feature dimensions of 8, 64, and 256, respectively, and starting from top the 1st, 2nd, and 3rd rows represent 1, 3, and 6 hidden layers, respectively.}
    \label{fig:appendix_l2_error}
\end{figure}

\begin{figure}[htbp]
    \centering
    \includegraphics[width=0.5\textwidth]{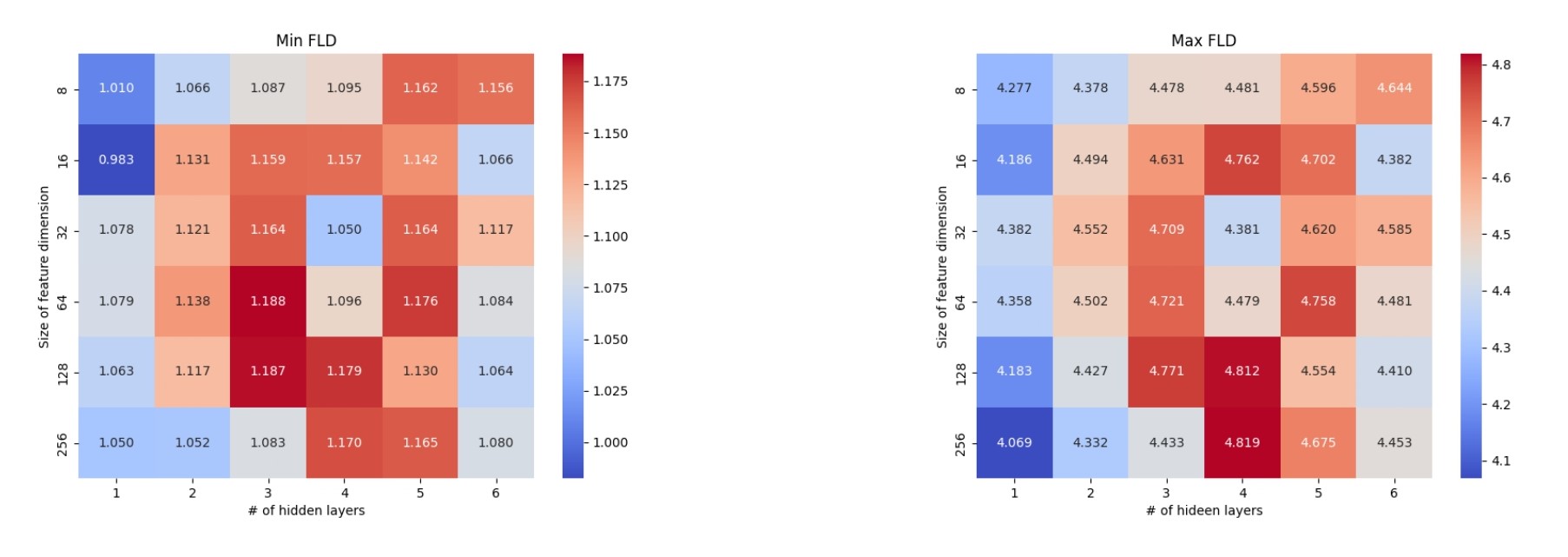}
    \caption{(Left) Minimum and (Right) Maximum of FLD between ASV and other 5 anomaly datasets (NET, ASVS, Coverdale, Geneva, and KJV Strongs). A higher minimum FLD indicates better differentiation between ASV and anomaly L2 error distributions.}
    \label{fig:appendix_fld}
\end{figure}

The L2 error distribution for each model is presented in Figure \ref{fig:appendix_l2_error}. The minimum Fisher’s Linear Discriminant (FLD) between the L2 norm distributions of the reconstructed sentence vectors from the trained dataset (ASV) and the anomaly datasets (NET, ASVS, Coverdale, Geneva, KJV Strongs) across the 36 models is shown in Figure \ref{fig:appendix_fld}.

The minimum FLD is more important than the maximum FLD for determining the separation between normal and anomaly data. A high minimum FLD represents the model that has the most differentiation between the ASV original and the anomaly reconstructions, indicating the best-performing model in terms of distinguishing between the original and anomalous styles based on the L2 norm distribution. Figure \ref{fig:appendix_fld} shows that the minimum FLD is maximized in models with 3 hidden layers and a feature dimension size between 32 and 128. Models with too small or too large hidden layers and feature dimensions tend to perform poorly in anomaly differentiation.

Across the 36 models, the anomaly dataset that produced the minimum FLD most frequently was Geneva, appearing 31 times, followed by KJV Strongs, which appeared 5 times. This suggests that the L2 error distribution of the Geneva dataset was generally the closest to that of ASV, making it the hardest to distinguish from ASV. Conversely, the anomaly dataset that consistently produced the maximum FLD in all 36 models was Coverdale, indicating that it was the easiest to distinguish from ASV based on the L2 error distribution. This result highlights the distinctiveness of Coverdale's style compared to ASV, while Geneva's style appears more similar.

\subsection{Impact of Context Subtraction on VAE Performance}

\begin{figure}[htbp]
    \centering
    \includegraphics[width=0.5\textwidth]{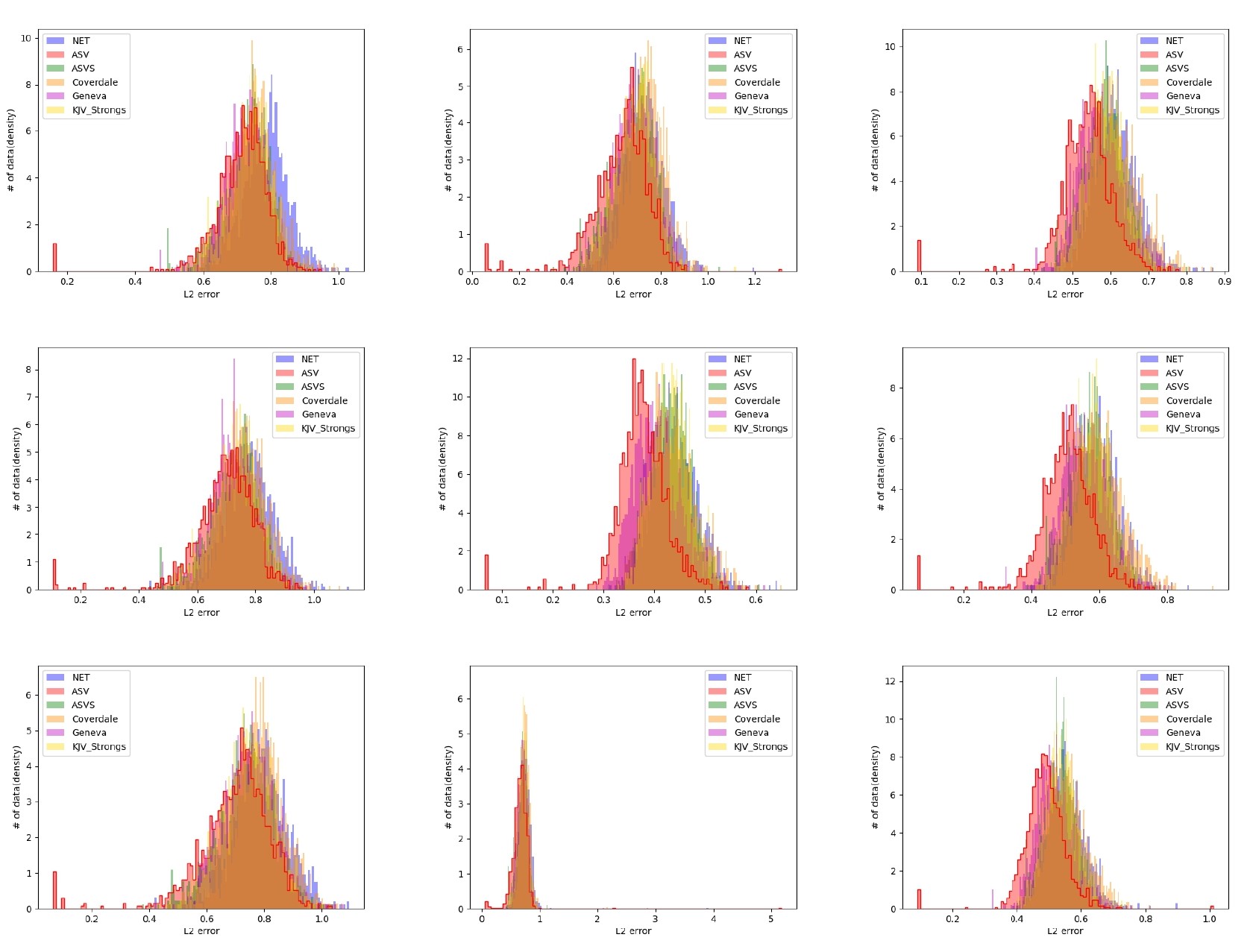}
    \caption{L2 error distribution on ASV, NET, ASVS, Coverdale, Geneva, and KJV Strongs, without parallel sentence (KJV) subtraction. The x-axis represents the L2 error between the original and reconstructed sentence vector, and the y-axis represents the distribution density. The hyperparameters of each model are as follows: starting from left the 1st, 2nd, and 3rd columns represent feature dimensions of 8, 64, and 256, respectively, and the starting from top 1st, 2nd, and 3rd rows represent 1, 3, and 6 hidden layers, respectively.}
    \label{fig:appendix_no_sub_l2_error}
\end{figure}

\begin{figure}[htbp]
    \centering
    \includegraphics[width=0.5\textwidth]{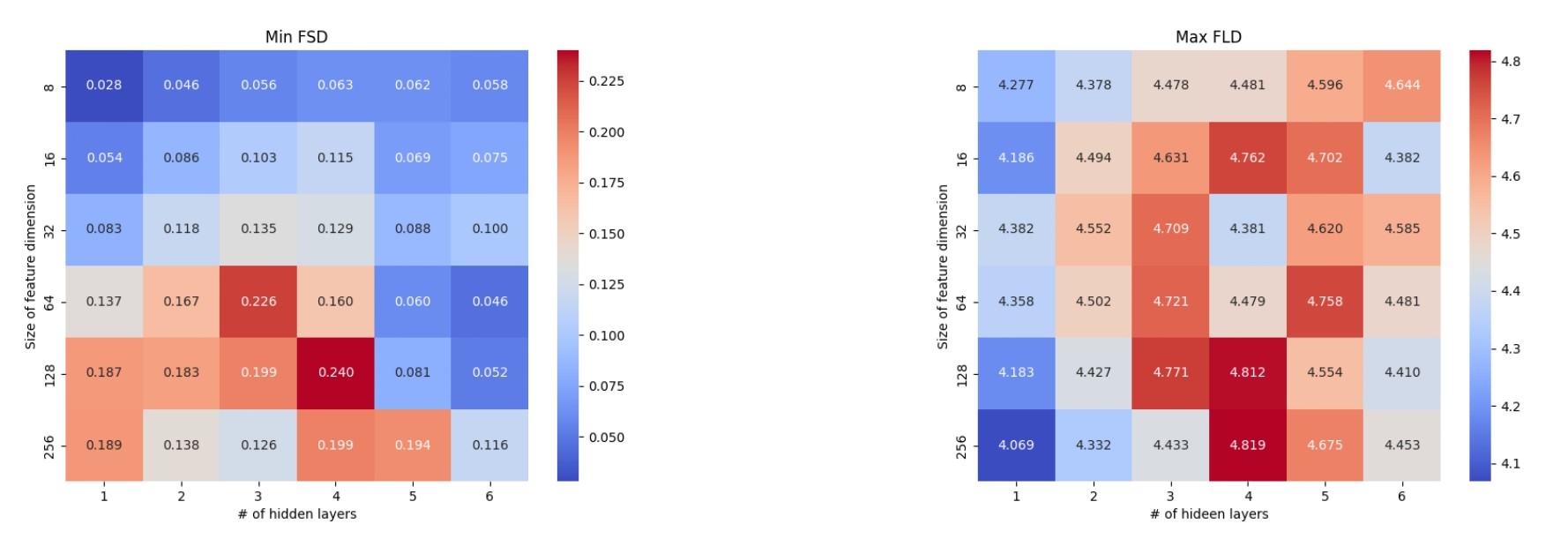}
    \caption{(Left) Minimum and (Right) Maximum of FLD between ASV and other 5 anomaly datasets (NET, ASVS, Coverdale, Geneva, and KJV Strongs), without parallel sentence (KJV) subtraction. A higher minimum FLD indicates better differentiation between ASV and anomaly L2 error distributions.}
    \label{fig:appendix_no_sub_fld}
\end{figure}

Training the VAE without subtracting context parallel sentence (KJV) vectors demonstrated that both the training loss and test loss decreased and converged, indicating successful learning. However, as shown in Figure \ref{fig:appendix_no_sub_l2_error} in Appendix F, the mean L2 error across all distributions was higher compared to the models trained with parallel sentence subtraction.

When comparing Figures \ref{fig:appendix_no_sub_fld} and \ref{fig:appendix_fld}, the Fisher’s Linear Discriminant (FLD) for the no-subtraction case (from context parallel sentence vectors) is significantly lower than for the subtracted case. Specifically, the mean of the minimum FLD across the 36 models in the subtracted case is 1.111, while the mean for the no-subtraction case is 0.116, making the FLD approximately 9.6 times lower without subtraction.

Furthermore, the highest maximum FLD in the no-subtraction case (1.000) is nearly the same as the lowest minimum FLD in the subtracted case (0.983). This stark difference in FLD highlights that when trained without subtracting the context parallel sentence vectors, the VAE's ability to distinguish anomalies from normal (trained domain) data is significantly diminished. This result reinforces the idea that the subtraction of context helps the VAE better capture stylistic differences, leading to clearer separation between ASV and other translations.

\section{Discussion}\label{sec:discussion}

This study extracted the styles of various Bible translations and utilized a Variational Autoencoder (VAE) model to analyze how these styles differ, particularly in comparison to the American Standard Version (ASV). The results revealed that the styles of each Bible translation followed a normal distribution, and these distributions could be clearly distinguished from that of the ASV. This indicates that there are stylistic differences between the ASV and other translations, and that these differences can be effectively detected using the VAE model.

After optimizing the VAE model’s hyperparameters, the process of distinguishing between the ASV and other translation styles resulted in a Type 1 error of 8.7\% and a Type 2 error of 6.7\%, with a total error rate of 15.3\%. Conversely, the model achieved an accuracy of 84.7\%, demonstrating its ability to effectively differentiate styles. This level of accuracy suggests that the model can clearly recognize the distribution of a specific style and use it as a basis to distinguish between the styles of different translations.

However, the VAE model was optimized for distinguishing a single style. While it was useful for detecting differences between a specific translation style and the ASV, it had limitations when it came to distinguishing multiple styles simultaneously or understanding the relationships between complex, multi-dimensional styles. These limitations stem from the structural characteristics of the VAE, which compresses the data’s features during learning, making it inherently challenging to fully capture the complex characteristics of the data. Therefore, to distinguish multiple styles simultaneously, it may be necessary to use other models or train the VAE model in a more sophisticated manner.

The ability to extract a specific style suggests that the style’s characteristics can be quantified and represented as a probability distribution. This means that AI can utilize this quantified style representation to generate text that adheres to a specific style. For example, in text generation tasks where a particular writing style or tone is required, a 'style metric' could be used as a numerical and comparable indicator to assess and ensure that the generated text conforms to the desired style.

The approach taken in this study opens up the possibility of expanding the research to other parallel text datasets. By applying this methodology to other text domains, researchers can study the stylistic differences and their implications within each domain. For example, the approach could be extended to analyze the styles of different translations of literary works, legal document translations, or works by various authors.

We have demonstrated that the VAE model can distinguish between the original and anomaly data using the reconstruction L2 error. To measure the overall accuracy, False Positive Rate (FPR), and False Negative Rate (FNR) of the model, we created an Accuracy Test Dataset using data not included in the training set. This dataset consisted of 1,000 samples, with 50\% of the samples being from ASV and the remaining 50\% from five anomaly datasets (NET, ASVS, Coverdale, Geneva, and KJV Strongs).

\begin{table}[t]
\caption{Accuracy \& Error Rates of Models 1, 2, and 3 on Anomaly Detection}
\label{tab:accuracy}
\vskip 0.15in
\begin{center}
\begin{small}
\begin{sc}
\setlength{\tabcolsep}{4pt} 
\begin{tabular}{|c|c|c|c|}
\hline
\textbf{Model} & \textbf{Accuracy} & \textbf{Type I Error} & \textbf{Type II Error} \\
\hline
Model 1 & 83.5\% & 9.8\% & 6.7\% \\
Model 2 & 82.9\% & 10.1\% & 7.0\% \\
Model 3 & 83.4\% & 9.8\% & 6.8\% \\
\hline
\textbf{Average} & \textbf{83.3\%} & \textbf{9.9\%} & \textbf{6.8\%} \\
\hline
\end{tabular}
\end{sc}
\end{small}
\end{center}
\vskip -0.1in
\end{table}

The binary classification results showed that the lowest overall error rate was achieved when the threshold was set at mean + 0.8 std. The average overall error rate across the three models was 16.8\%. The relatively high FNR, particularly with Geneva, suggests that modern English Bible translations inherently do not exhibit distinct stylistic differences.

The results of the anomaly detection using the VAE in this study also show trends similar to what would be expected when humans classify ASV and other Bible versions. In this study, the L2 error distributions of ASV and Geneva had a significant overlap, making it difficult to classify them with a low error rate using a specific threshold. In contrast, the L2 error distribution of Coverdale barely overlapped with ASV, and the FLD was the highest across all models. More typically, 67.8\% (ASV 34.6\%, KJV Strongs 33.2\%) of type 2 error in anomaly detection is from Geneva and KJV Strongs.

\begin{table}[t]
\caption{Original Sentences of 3 Different Versions: ASV, Geneva, Coverdale}
\label{tab:verses}
\vskip 0.15in
\begin{center}
\begin{small}
\begin{sc}
\begin{tabularx}{0.45\textwidth}{lX}
\toprule
\textbf{Verse} & \textbf{Translation} \\
\midrule
\textbf{Gen 1:1} & \textbf{ASV:} In the beginning God created the heavens and the earth. \newline
\textbf{Geneva:} In the beginning God created the heauen and the earth. \newline
\textbf{Coverdale:} In ye begynnynge God created heauen \& earth: \\
\midrule
\textbf{Mat 1:1} & \textbf{ASV:} The book of the generation of Jesus Christ, the son of David, the son of Abraham. \newline
\textbf{Geneva:} The book of the generation of Jesus Christ the son of David, the son of Abraham. \newline
\textbf{Coverdale:} This is the boke of the generacion of Iesus Christ ye sonne of Dauid, the sonne of Abraham. \\
\midrule
\textbf{Joh 3:16} & \textbf{ASV:} For God so loved the world, that he gave his only begotten Son, that whosoever believeth on him should not perish, but have eternal life. \newline
\textbf{Geneva:} For God so loveth the world, that he hath given his only begotten Son, that whosoever believeth in him, should not perish, but have everlasting life. \newline
\textbf{Coverdale:} For God so loued the worlde, that he gaue his onely sonne, that who so euer beleueth in hi, shulde not perishe, but haue euerlastinge life. \\
\bottomrule
\end{tabularx}
\end{sc}
\end{small}
\end{center}
\vskip -0.1in
\end{table}

Table \ref{tab:verses} illustrates the textual differences between three versions of the Bible (ASV, Geneva, Coverdale), which could influence the VAE's ability to distinguish anomalies. The relatively low accuracy of anomaly detection using the VAE in this study may be attributed to the subtle stylistic differences between the texts. This implies that using sentences with clearer stylistic differences and more varied contexts in future experiments could result in better accuracy.

%% file: 5.Conclusion.tex
\section{Conclusion}\label{sec:conclusion}

This study has successfully demonstrated the application of a Variational Autoencoder (VAE) model to analyze and distinguish the stylistic differences among various Bible translations, with a particular focus on the American Standard Version (ASV). By embedding textual data into high-dimensional vectors and applying anomaly detection techniques, the study identified unique stylistic distributions for each translation, showcasing the model’s capability to differentiate between these styles with a notable accuracy rate of 84.7\%. The findings confirm that the VAE model can effectively capture and differentiate textual styles, though it is primarily optimized for distinguishing a single style.

Despite these successes, the study also highlighted certain limitations inherent in the VAE model, particularly its challenges in simultaneously distinguishing multiple styles or comprehending complex, multi-dimensional stylistic relationships. This limitation underscores the need for further refinement of the model, potentially through the integration of more advanced machine learning techniques, such as deep learning models capable of classification or other unsupervised learning methods. Addressing these limitations could enhance the model’s ability to manage more complex stylistic differentiation tasks.

The implications of this research extend beyond academic inquiry, offering significant potential applications in the field of AI-driven text generation. The ability to extract and measure specific stylistic features opens up possibilities for generating texts with targeted stylistic attributes, which can be invaluable in automated writing tools, personalized content creation, and stylistic analysis of literary works. Moreover, the methodology employed in this study can be adapted to other text domains, providing a framework for analyzing stylistic differences across various types of textual data, including literary translations, legal documents, and author-specific writing styles.

In conclusion, while this study provides a solid foundation for the analysis of textual styles using VAE, it also sets the stage for future research to explore more sophisticated models and methodologies. By expanding this approach to other text domains and enhancing the model’s capabilities, future work can continue to deepen our understanding of textual styles and their applications in AI and beyond.